\pdfoutput=1

\documentclass[11pt]{article}

\usepackage[final]{acl}

\usepackage{times}
\usepackage{latexsym}

\usepackage[T1]{fontenc}

\usepackage[utf8]{inputenc}

\usepackage{microtype}

\usepackage{inconsolata}

\usepackage{amsmath,amsfonts,bm}

\def\eqref#1{equation~\ref{#1}}

\def\1{\bm{1}}

\DeclareMathAlphabet{\mathsfit}{\encodingdefault}{\sfdefault}{m}{sl}
\SetMathAlphabet{\mathsfit}{bold}{\encodingdefault}{\sfdefault}{bx}{n}

\newcommand{\KL}{D_{\mathrm{KL}}}

\usepackage{hyperref}
\usepackage{url}
\usepackage[pdftex]{graphicx}
\usepackage{subfigure}
\usepackage{subcaption}
\usepackage{tabularx}
\usepackage{multirow}
\usepackage{adjustbox}

\usepackage{booktabs}
\usepackage{mathtools}
\usepackage{longtable}
\usepackage{makecell}
\usepackage{hhline}
\usepackage{booktabs}
\usepackage{tablefootnote}
\usepackage{threeparttable}
\usepackage{wrapfig}
\usepackage{enumitem}
\usepackage[normalem]{ulem}
\usepackage{stfloats}
\usepackage{setspace}

\title{
LoRA Meets Dropout under a Unified Framework
}

\author{
\bf Sheng Wang$^{\heartsuit}$\thanks{\ \ Equal Contribution.}~,
Liheng Chen$^{\heartsuit \ast}$,
Jiyue Jiang$^{\spadesuit}$,
Boyang Xue$^{\spadesuit}$, \\
\bf Lingpeng Kong$^{\heartsuit}$,
Chuan Wu$^{\heartsuit}$ \\
$^{\heartsuit}$ The University of Hong Kong, $^{\spadesuit}$ The Chinese University of Hong Kong \\
{\tt
\{u3009618, clh648\}@connect.hku.hk,
jiangjy@link.cuhk.edu.hk, 
} \\
{\tt
byxue@se.cuhk.edu.hk,
\{lpk, cwu\}@cs.hku.hk
}
}

\begin{document}

\maketitle
\begin{abstract}
With the remarkable capabilities, large language models (LLMs) have emerged as essential elements in numerous NLP applications, while parameter-efficient finetuning, especially LoRA, has gained popularity as a lightweight approach for model customization.
Meanwhile, various dropout 
methods, initially designed for full finetuning with all the parameters updated, alleviates overfitting associated with excessive parameter redundancy. Hence, a possible contradiction
arises from negligible trainable parameters of LoRA and the effectiveness of previous dropout methods, which has been largely overlooked.
To fill this gap, we first confirm that parameter-efficient LoRA is also overfitting-prone. We then revisit transformer-specific dropout methods, and establish their equivalence and distinctions mathematically and empirically. Building upon this comparative analysis, we introduce a unified framework for a comprehensive investigation, which instantiates these methods based on dropping position, structural pattern and compensation measure. 
Through this framework, we reveal the new preferences and performance comparisons of them when involved with limited trainable parameters. This framework also allows us to amalgamate the most favorable aspects into a novel dropout method named HiddenKey. Extensive experiments verify the remarkable superiority and sufficiency of HiddenKey across multiple models and 
tasks, which highlights it as the preferred approach for high-performance and parameter-efficient finetuning of LLMs.


\end{abstract}

\section{Introduction}

Recently, transformers~\citep{Vaswani2017}, such as GPT-4~\citep{OpenAI2023}, PaLM 2~\citep{Anil2023} and LLaMA 2~\citep{Touvron2023a},
have been rapidly expanded to 
billions of parameters, leading to remarkable performance boost.
When customizing these models for downstream tasks, parameter-efficient fine-tuning (PEFT)  \citep{Houlsby2019, Hu2021, Liu2022} has been widely adopted as a lightweight method, which generally freezes the majority of parameters while only updating or adding negligible trainable parameters. Among these methods, LoRA \citep{Hu2021} gains the most popularity
due to its high effectiveness, robustness and generality. 

In parallel with this, 
dropout~\citep{Hinton2012} has been widely adopted to mitigate overfitting, which is generally caused by excessive parameter redundancy.
Its variants, including DropKey \citep{Li2023}, DropAttention \citep{Zehui2019} and HiddenCut \citep{Chen2021}, have also
demonstrated 
superiority
for transformers.
With a specified probability, they randomly deactivate attention logits, weights and hidden representations, respectively.
However, the effectiveness of these methods is only verified in full finetuning scenarios, where all the parameters are updated and easily lead to excessive redundancy.
When it comes to LoRA-based PEFT scenarios, a potential contradiction arises. Specifically, \textit{since overfitting primarily stems from excessive parameter redundancy, dropout may prove ineffective in LoRA-based finetuning because of the extremely limited trainable parameters.} Besides, all the above methods are proposed independently, lacking a clear guideline to unify them systematically, which hinders comprehensive comparative analysis and the development of more effective dropout methods.

In this study, we first conduct extensive experiments and confirm that LoRA also suffers from overfitting easily, which serves as a prerequisite for our following analysis.
As shown in Figure~\ref{fig:dropout}, as the rank and trainable parameters increase, 
the model's performance initially improves but gradually deteriorates due to the intensifying overfitting.
Much more experiments in Sec.~\ref{sec:exp} provide further evidence and affirm that this overfitting susceptibility can be improved with dropout methods.
Besides, we compare the above transformer-specific dropout methods mathematically and empirically. For the first time, we find that DropKey and DropAttention share the equivalent forwarding process, while the gradient stopping operator introduces gradient noise into the backpropagation of DropAttention, impairing the training stability and performance. 

Based on the comparative analysis, we identify three key dimensions for a dropout method and derive a unified framework along
dropping position, structural pattern and compensation measure.
With this framework, empirical experiments firstly reveal the new preferences of these methods in LoRA scenarios. For example, span-wise HiddenCut is no longer superior to the element-wise one due to the limited tunable parameters.
Secondly, this framework enables the comprehensive comparisons among different methods. 
Empirically, we find that DropKey performs the best followed by HiddenCut, 
while DropAttention exhibits the worst performance due to the gradient noise. As an alternative compensation, Bidirectional Kullback-Leibler (KL) divergence loss consistently achieves performance gains, while Jensen-Shannon (JS) consistency regularization loss becomes ineffective.

Guided by this framework, we also derive a new dropout method named HiddenKey,
which drops attention logits column-wisely and hidden representations element-wisely, respectively, and augment the vanilla loss with KL loss. 
It consistently exhibits superiority across multiple models in both natural language understanding (NLU) and generation (NLG) tasks, which also fills the largely overlooked gap on the effect of dropout methods on NLG tasks.
Integrating with input and output dropout does not provide further complementarity, demonstrating the sufficiency of our method.
Hence, HiddenKey excels as the better method for high-performance and parameter-efficient finetuning of LLMs on both NLU and NLG tasks.

In summary, our contributions are mainly as follows:
\begin{itemize}[]
  \item We present the first comprehensive investigation to explore the potential contradiction between various dropout methods and LoRA.

  \item We compare three typical transformer-specific dropout methods theoretically and empirically, and derive the core dimensions for designing a dropout method.
  
  
  \item We further introduce a unified framework to instantiate existing dropout methods, within which we discover the new preferences and performance comparison of these methods.
  
  \item A new dropout method named HiddenKey is devised within 
  our framework, exhibiting superior effectiveness and sufficiency in mitigating LoRA's susceptibility to overfitting.



\end{itemize}

\section{Preliminaries}

\begin{figure}[!ht]
   \centering
    \includegraphics[width=0.85\linewidth]{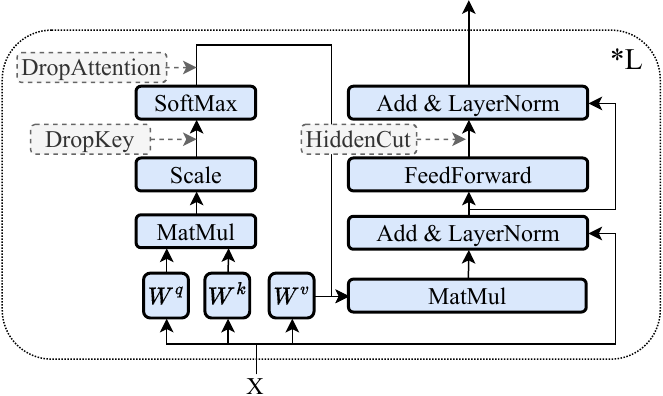}
    \caption{Illustration of transformer architecture and typical transformer-specific dropout methods, namely DropKey, DropAttention, and HiddenCut. 
    }
    \label{fig:transformer}
\end{figure}

In this section, we revisit three transformer-specific dropout methods shown in Figure~\ref{fig:transformer}, laying the foundation for the subsequent analysis.
\paragraph{DropAttention.}
DropAttention~\citep{Zehui2019} is the first dropout method specially designed for self-attention mechanism. It randomly masks elements or key columns of attention weights, encouraging the utilization of multiple contextualized features instead of overfitting some specific patterns. Following Eq.~\ref{eq:DA} and \ref{eq:norm rescale}, normalized rescaling replaces the traditional one to guarantee the sum of attention weights to be one, and achieves better performance
for multiple NLP classification tasks.

\vspace{-0.8\baselineskip}
\begin{small}
\begin{align}
    \overline{w}_j & = m \cdot w_j, \quad m \sim \operatorname{Bernoulli}(p) , \label{eq:DA} \\
    w_j^\prime & = \frac{\overline{w}_j}{ \operatorname{NoGrad}(\sum_{j=0}^{l-1} \overline{w}_{j})}, \label{eq:norm rescale}
\end{align}
\end{small}
where $p$, $l$, ${w}_j$, $\overline{w}_j$, and $w_j^\prime$ denote the dropout rate, sequence length, original, masked, and rescaled attention weights. $\operatorname{NoGrad}()$ and $\operatorname{Bernoulli}()$ represent the gradient stopping operator and sampling from the Bernoulli distribution, respectively\footnote{Here we omit the subscript $t$ for clarity. Although whether the $\operatorname{NoGrad}()$ operator exists or not 
significantly impacts 
the performance of DropAttention, it is overlooked in the original paper. We present it here and will discuss both cases in detail.}.  

\paragraph{DropKey.}
As a dropout-before-softmax scheme, DropKey \citep{Li2023} takes attention logits $g_j$ instead of weights as the basic units, as formulated in Eq.~\ref{eq:DK mask}. Since 
the subsequent
$\operatorname{softmax}()$ ensures the sum of weights to be one, rescaling is no longer necessary. 

\vspace{-\baselineskip}
\begin{small}
\begin{align}
    g_j^\prime = m + g_j, ~~
    m =
    \begin{cases}
        0, & \text{with probability } 1-p \\
        -\infty, & \text{with probability } p
    \end{cases}
    \label{eq:DK mask}
\end{align}
\end{small}

\paragraph{HiddenCut.} 
In contrast, HiddenCut \citep{Chen2021} 
focuses on preventing the co-adaptation
of hidden representations in the feed-forward module.
The core idea is to cut single contiguous span, which may contain more semantic information and be more difficult to be restored.
Besides, JS loss is applied to encourage
the perturbed representations to be as close to those in inference as possible.


\begin{figure}[!ht]
        \centering
        \subfigure[element]{
            \includegraphics[width=0.25\linewidth]{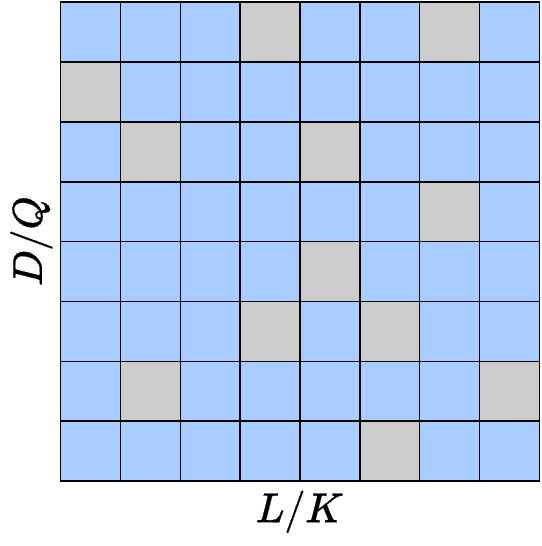}
            \label{fig:element}
            }
        \subfigure[column]{
            \includegraphics[width=0.25\linewidth]{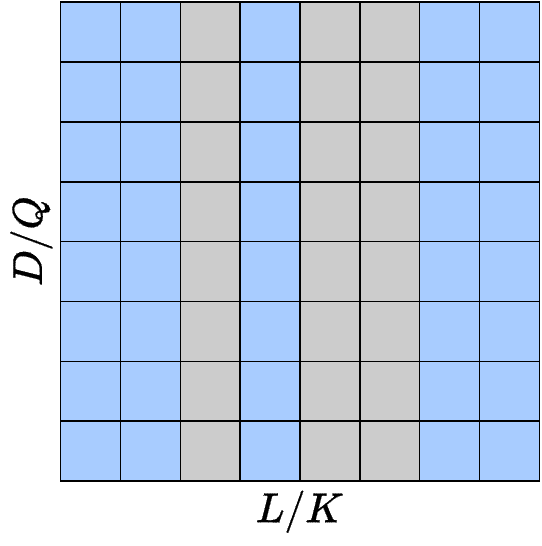}
            \label{fig:column}
            }
        \subfigure[span]{
            \includegraphics[width=0.25\linewidth]{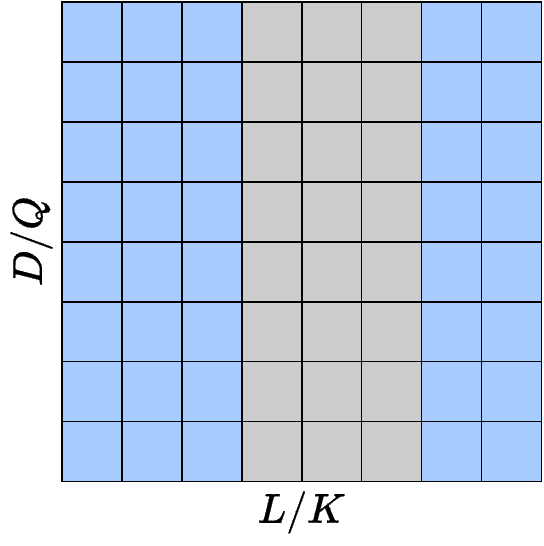}
            \label{fig:span}
            }
        \caption{Three structural sampling strategies, namely element, column, and span. The grey and blue cells represent masked and remaining entries, respectively. 
        In HiddenCut, rows and columns denote sequence length ($L$) and hidden dimension ($D$), while representing keys ($K$) and queries ($Q$) in DropKey and DropAttention.
        }
        \label{fig:structured dropout}
\end{figure}

\section{Method}
Firstly, we conduct a comparative analysis of the above methods. Based on their similarity and differences, we then propose a unified framework along dropping position, structural pattern and compensation measure.
Finally, this framework guides us to derive a new dropout method named HiddenKey, which 
exhibits superior performance empirically.


\subsection{Mathematical and Empirical Comparison}  \label{sec:math}
\paragraph{Equivalent Forwarding between DropKey and DropAttention.} Despite the different details between DropKey and DropAttention, we show their mathematical equivalence in forwarding.
Let $g_u$ and $g_m$ denote the unmasked and masked attention logits, while $w_u$ and $w_m$ represent the corresponding attention weights\footnote{Only one masked element is considered here, but masking multiple elements shares the same analysis.
}. For DropKey, we have

\vspace{-1.2\baselineskip}
\begin{small}
\begin{align}
    g_m^\prime & \coloneqq -\infty, \quad g_u^\prime \coloneqq g_u, \quad w_m^\prime=0 , \\
    w_u^\prime & = \frac{\exp (g_u^\prime)}{\sum_{i=0}^{l-1} \exp (g_i^\prime)} ,  \label{eq:dkwu}
\end{align}
\end{small} \vspace{-0.6\baselineskip} \\ 
while for DropAttention, we have \vspace{-\baselineskip} \\ 
\begin{small}
\begin{align}
w_m^\prime  \coloneqq 0 , \quad
w_u^\prime  = \frac{\exp (g_u)}{\sum_{i=0}^{l-1} \exp (g_i)} \cdot \frac{1}{\sum_{i=0}^{l-1}{\overline{w}_i}} . \label{eq:dawu}
\end{align}
\end{small}  \\
Proved by Eq.~\ref{eq:equivalence} in Appendix~\ref{apdx: math}, Eq.~\ref{eq:dkwu} and Eq.~\ref{eq:dawu} are strictly equal to each other. Hence, the final attention weights (i.e., $w_u^\prime$ and $w_m^\prime$) of DropKey are the same as those of DropAttention,
and so is the following computation. Notably, 
normalized rescaling plays an indispensable role in establishing this equivalence, which diminishes the differences between these two methods during the forward pass.

\paragraph{Variation in Back-Propagation between DropKey and DropAttention.}
Due to the equivalent forward pass,  
the corresponding 
values of 
$\frac{\partial O}{\partial w_u^\prime} $ and $\frac{\partial O}{\partial w_m^\prime} $ remain the same for DropKey and DropAttention, where $O$ denotes the objective function. Meanwhile, because of the identical  computation before attention logits, the analysis of back-propagation focuses on the four partial derivatives of $w_u^\prime$ and $w_m^\prime$ with respect to $g_u$ and $g_m$, respectively. For DropKey, we have 

\vspace{-\baselineskip}
\begin{small}
\begin{align}
    \frac{\partial w_u^\prime}{\partial g_u} & = \exp(g_u) \cdot \frac{\sum_{i=0, \neq m}^{l-1} \exp (g_i) - \exp(g_u)}{ (\sum_{i=0, \neq m}^{l-1} \exp (g_i))^2}.
\end{align}
\end{small} 
For DropAttention with $\operatorname{NoGrad}()$, we have
\begin{small}
\begin{align}
    \frac{\partial w_u^\prime}{\partial g_m} & = - \frac{\exp(g_u) \cdot \exp(g_m)}{ \sum_{i=0}^{l-1} \exp (g_i) \cdot \sum_{i=0, \neq m}^{l-1} \exp (g_i)} , \\
    \frac{\partial w_u^\prime}{\partial g_u} & =  \frac{\exp(g_u) \cdot \sum_{i=0, \neq u}^{l-1} \exp (g_i)}{ \sum_{i=0}^{l-1} \exp (g_i) \cdot \sum_{i=0, \neq m}^{l-1} \exp (g_i)} . 
\end{align}
\end{small} 
As for other partial derivatives, their gradient flow is disrupted by dropping operations.
When the corresponding elements of attention logits and weights are masked in DropKey and DropAttention,
the derivative of $w_u^\prime$ with respect to $g_u$ has proportional relation, as shown in Eq.~\ref{eq:propotion} and proven by Eq.~\ref{eq:k prove}. 
Provably, $k$ is always less than 1 and continuously decreases with the increase of $g_m$. 
In other words, compared to DropAttention with $\operatorname{NoGrad}()$,
DropKey can adaptively lower the gradients when a large attention logit $g_m$ is discarded. This can provide DropKey with dropping-dependent compensation capability, thereby stabilizing the training process. 
For DropAttention with $\operatorname{NoGrad}()$, the partial derivative of $w_u^\prime$ with respect to $g_m$ is non-zero and that with respect to $g_u$ depends on the value of $g_m$, even if $w_m$ is masked and $g_m$
is not used for 
computation. This implies that a larger dropout rate can introduce more gradient noise, which is further validated by the inferior performance in Sec.~\ref{sec:exp}.
In contrast, DropAttention without $\operatorname{NoGrad}()$ shares the same back-propagation with DropKey, thereby exhibiting identical behaviors. Hence, unless otherwise stated, we will refer to DropAttention with $\operatorname{NoGrad}()$ as DropAttention, and include DropAttention without $\operatorname{NoGrad}()$ under DropKey for simplicity.

\vspace{-\baselineskip}
\begin{small}
\begin{align}
(\frac{\partial w_u^\prime}{\partial g_u} )^{\text{DropKey}} &= k \cdot  (\frac{\partial w_u^\prime}{\partial g_u} )^{\text{DropAttention}} , \label{eq:propotion} \\
k&=\frac{1-\frac{\exp(g_u)}{\sum_{i=0, \neq m}^{l-1} \exp (g_i)}}{1-\frac{\exp(g_u)}{\sum_{i=0}^{l-1} \exp (g_i)}} \nonumber 
\end{align}
\end{small}
\vspace{-\baselineskip}
\paragraph{Comparison with HiddenCut.}
The commonality among these methods is that they all need to select a specific type of data, decide what patterns to mask, and consider how to reduce the gap between training and inference phases.
In contrast, their divergences are two-fold. 
First, their distinct dropping positions and patterns lead to different rescaling operators. Identical to the vanilla dropout, element-wise HiddenCut amplifies hidden representations by 
a factor of 
$1/(1-p)$ for consistent scales between training and testing,
while normalized rescaling is adopted by DropAttention. Due to the subsequent $\operatorname{softmax}()$, DropKey no longer utilizes any rescaling method.
The other difference is that 
DropAttention and DropKey can be regarded as operations on weight matrices, which are utilized for the weighted summation of value vectors.
Instead, HiddenCut operates directly in the hidden representations.

In summary, the comparative analysis of these methods highlights their similarities and differences, leading to the identification of key dimensions for designing a dropout method: dropping position, structural pattern and compensation measure. Subsequently, these elements are incorporated into our unified framework for further analysis.

\subsection{A Unified Framework}
Based on the above comparative analysis, we identify three key dimensions for a dropout method. Here we elaborate them further and instantiate these dropout methods along them below.

\paragraph{Dropping Position.}
For better generalization,
a robust model needs to learn noise-resilient features.
Hence, dropping position, determining where to inject noise, 
emerges as a primary consideration in designing dropout methods. For example, dropping inputs acts like data augmentation, dropping outputs
encourages an ensemble of sub-classifiers, and dropping intermediate representations disrupts the co-adaptation of neighboring neurons. For a transformer layer depicted in Figure~\ref{fig:transformer}, DropKey, DropAttention and HiddenCut respectively drop attention logits, weights and hidden representations, covering the self-attention mechanism and feed-forward module. 
Additionally, the same dropping position may perform differently in full finetuning and LoRA scenarios.
In full finetuning, weights located in the dropping position are directly adjusted for better noise resilience. However, this adaptation is more implicit for LoRA, because the directly associated weights with the dropping position are frozen. Specifically, LoRA, typically applied to the key and value projection matrices~\citep{Hu2021}, requires multiple intermediate calculations (e.g., softmax) to influence attention logits and weights (i.e., the dropping positions for DropAttention and DropKey), while even requires inter-module computation for hidden representations. This disparity may potentially affect the effectiveness of existing dropout methods in LoRA scenarios.
Notably, distinct dropping positions do not necessarily indicate differences. 
In specific cases, different positions may also exhibit similar features, as discussed in Sec.~\ref{sec:math}.

\paragraph{Structural Pattern.}
Structural pattern means the style of units deactivated randomly, and determines how the co-adaptation of neurons is disrupted, thereby 
affecting the semantic information learned by these units. For example, as shown in Figure~\ref{fig:column}, if column pattern is adopted in DropKey, each value vector tend to possess as much contextual information as possible so that the output vectors are minimally affected by the masked key columns.
Different patterns also result in varying levels of difficulty in recovery~\citep{Zehui2019}. 
Generally, the span pattern is more challenging than the column style, while the element one is the simplest.
Given the limited trainable parameters,
LoRA may struggle to handle the strong disturbances introduced by complex patterns. Therefore, it may exhibit different preferences for structural patterns from full finetuning.
Besides, different optimal patterns may be required for distinct positions, 
which will be thoroughly discussed in Sec.~\ref{sec:exp}.

\paragraph{Compensation for Training and Inference Gap.}
For better performance and deterministic outputs, dropout 
is 
disabled in inference by default. However, this is not consistent with the training stage and can lead to a gap between the actual and ideal performance. Hence, another key consideration is how to close the training and inference gap. 
Apart from rescaling associated with each method intrinsically, R-drop \citep{Wu2021}
leverages Eq.~\ref{eq:kl}, bidirectional KL divergence loss, to enforce the output distributions to be more dropout-insensitive so that the gap
can be minimized implicitly.
Alternatively, HiddenCut replaces it
with JS loss shown in Eq.~\ref{eq:js}. 
With negligible tunable parameters, LoRA is more easily optimized to reach its performance ceiling. This compressed optimization space may potentially render some existing schemes ineffective, which is also verified in the following sections. 

\vspace{-\baselineskip}
\begin{small}
\begin{align}
\mathcal{L}_{KL} & = \frac{1}{2}(\KL(P_1 \Vert P_2)+\KL(P_2 \Vert P_1)) \label{eq:kl} , \\
\mathcal{L}_{JS} & = \KL(P_1 \Vert \overline{P}) , \label{eq:js}
\end{align}
\end{small} \vspace{-\baselineskip} \\ 
where $P_1$, $P_2$, and $\overline{P}$ represent two different output distributions in the training stage and one in inference with the same input, respectively. For the sake of symmetry, KL loss calculates the bidirectional distances, while JS loss uses the inference distribution as reference.

\subsection{HiddenKey}
\begin{figure*}[!ht]
    \centering
    \includegraphics[width=0.9\linewidth]{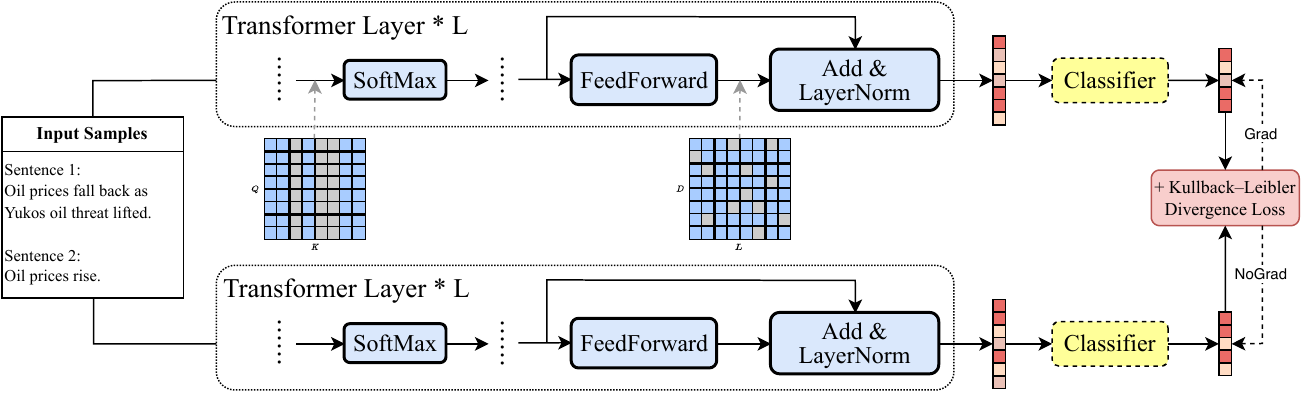}
    \caption{Illustration of HiddenKey. It respectively drops columns and elements of attention logits and hidden representations, and augments bidirectional KL loss to minimize the training and inference gap implicitly.
    %
    }
    \label{fig:siamese}
\end{figure*}
The proposed unified framework not only enables us to analyze the critical choices along each dimension and their mutual influences, but also guides us to design new dropout methods. As shown in Figure~\ref{fig:siamese}, we propose ``HiddenKey'', which drops the attention logits column-wisely in the attention mechanism and hidden representations element-wisely in the feed-forward module along the dropping position and structural pattern dimensions. As for the compensation measure to minimize the training and inference gap, two forward passes in parallel are performed so that an extra KL loss is deployed to enhance the similarity of output distributions.
For classification tasks, the representations produced by the classifier are used, while those produced by the last transformer layer are used for regression tasks. Furthermore, the superiority over all the aforementioned methods will be extensively analyzed on diverse tasks and models below.


\section{Experiments} \label{sec:exp}
\subsection{General Setup} \label{sec: general setup}
\paragraph{Models and Datasets.}
We implement comprehensive analysis on multiple tasks and models with LoRA.
The models start from RoBERTa-large~\citep{Liu_Ott_Goyal_Du_Joshi_Chen_Levy_Lewis_Zettlemoyer_Stoyanov_2019} and GPT2-Medium~\citep{Li2021}, and scale up to LLaMA2-7B~\citep{Touvron2023}.
Besides, both NLU and NLG tasks are covered. 
For NLU tasks, we utilize six datasets from GLUE benchmark~\citep{Wang2018}: 
\textbf{SST-2}~\citep{Socher_Perelygin_Wu_Chuang_Manning_Ng_Potts_2013},
\textbf{RTE}~\citep{Wang2018},
\textbf{MRPC}~\citep{Dolan_Brockett_2005},
\textbf{STS-B}~\citep{Cer2017},
\textbf{CoLA}~\citep{Warstadt2018},
and \textbf{QNLI}~\citep{Rajpurkar2018}. 
These datasets are selected to cover diverse tasks and sizes, including single sentence, similarity, paraphrase and inference. 
For NLG tasks, we follow \citet{Hu2021} and focus on \textbf{E2E} \citep{Novikova2017} and \textbf{WebNLG} \citep{Gardent_Shimorina_Narayan_Perez-Beltrachini_2017}. More details can be found in Appendix~\ref{apdx: dataset}.

\paragraph{Baseline.}
Due to the widespread popularity, we use vanilla LoRA as the baseline, and keep most of its  configurations.
Notably, low-rank decomposition with a rank of 8 and scalar of 16 is applied to the key and value projection matrices.
This results in trainable parameters of 0.79M in the Roberta-large model, accounting for 0.22\% of the total parameters\footnote{The classifier parameters are excluded here due to their varying numbers for different tasks.}.
In comparison, these values are 0.39M and 0.11\% for GPT2-Medium, 
while 4.19M and 0.06\% for LLaMA2-7B. 
More detailed configurations are demonstrated in the Appendix~\ref{apdx: config}.

\subsection{Main Results}

\begin{table*}[!ht]
    \centering
    \begin{tabular}{ccccccc}

  \toprule
  \multirow{2}{*}{Position}      & \multirow{2}{*}{\makecell{Pattern /\\Compen. }} & RTE                                   & MRPC                                    & STS-B                                    & STS2                                    &  \multirow{2}{*}{Avg.}         \\
  \hhline{|~~----~|}
                                             &               & Acc.                                 & Acc.                                      & Pearson.                      & Acc.                                    &                  \\
  \hline
  $\text{Full Finetuning}^{*}$               & -             & $\text{86.60}$ & $\text{90.90}$                         & $\text{ 92.40}$                           & $\text{96.40}$          & \text{91.58}     \\
  \hline
  Baseline                                  & -             & $\text{84.48}_{\pm 0.98}$      & $\text{89.95}_{\pm 0.50}$        & $\text{91.96}_{\pm 0.48}$         & $\text{95.99}_{\pm 0.25}$        &   90.60                                     \\
  \hline
  \multirow{3}{*}{HiddenCut}                & element       & $\text{87.00}_{\pm 1.14}$      & $\text{90.69}_{\pm 0.42}$        & $\text{91.94}_{\pm 0.28}$         & $\text{96.10}_{\pm 0.42}$        &   91.43                                    \\
                                            & column        & $\text{86.64}_{\pm 0.80}$      & $\text{90.20}_{\pm 0.80}$        & $\text{91.96}_{\pm 0.11}$         & $\text{96.22}_{\pm 0.19}$        &   91.26                                     \\
                                            & span          & $\text{86.64}_{\pm 1.63}$      & $\text{90.69}_{\pm 0.22}$        & $\text{92.05}_{\pm 0.35}$         & $\text{96.10}_{\pm 0.30}$        &   91.37                                     \\
  \hline
  \multirow{3}{*}{DropKey}                  & element       & $\text{87.00}_{\pm 1.08}$      & $\text{90.93}_{\pm 1.06}$        & $\text{92.21}_{\pm 0.21}$         & $\text{96.22}_{\pm 0.25}$        &   91.59                                  \\
                                            & column        & $\text{87.36}_{\pm 1.70}$      & $\text{90.93}_{\pm 0.40}$        & $\text{92.25}_{\pm 0.13}$         & $\text{96.22}_{\pm 0.24}$        &   91.69                                      \\
                                            & span          & $\text{86.28}_{\pm 0.94}$      & $\text{90.69}_{\pm 0.69}$        & $\text{92.21}_{\pm 0.21}$         & $\text{96.22}_{\pm 0.25}$        &   91.35                                    \\
  \hline
  \multirow{3}{*}{DropAttention }           & element       & $\text{85.56}_{\pm 11.73}$     & $\text{90.20}_{\pm 3.07}$        & $\text{92.03}_{\pm 0.27}$         & $\text{95.76}_{\pm 0.30}$        &   90.89                                    \\
                                            & column        & $\text{85.56}_{\pm 1.80}$      & $\text{90.20}_{\pm 0.71}$        & $\text{92.11}_{\pm 0.28}$         & $\text{95.87}_{\pm 0.21}$        &   90.94                                   \\
                                            & span          & $\text{86.28}_{\pm 0.60}$      & $\text{89.95}_{\pm 0.61}$        & $\text{92.21}_{\pm 0.36}$         & $\text{96.10}_{\pm 0.39}$        &   91.14                                   \\
  \hline
  \multirow{3}{*}{\makecell{$\text{HiddenKey}^{-}$ }} & -    & $\text{87.70}_{\pm 0.91}$ & $\text{90.90}_{\pm 0.72}$  & $\text{92.28}_{\pm 0.19}$   & $\text{96.22}_{\pm 0.13}$   &   91.78                                        \\
                                            & + KL          & $\text{88.10}_{\pm 1.60}$  & $\text{\bf 91.20}_{\pm 0.90}$        & $\text{\bf 92.30}_{\pm 0.11}$         & $\text{\bf 96.44}_{\pm 0.20}$        &   {\bf 92.01}                                   \\
                                            & + JS          & $\text{87.70}_{\pm 1.72}$      & $\text{90.90}_{\pm 0.47}$        & $\text{92.24}_{\pm 0.21}$         & $\text{96.22}_{\pm 0.24}$         &   91.77                                 \\
\hhline{|~------|}
             \quad ~+ input                               & -       & $\text{\bf 88.50}_{\pm 2.11}$      & $\text{90.70}_{\pm 1.03}$        & $\text{92.11}_{\pm 0.14}$         & $\text{96.33}_{\pm 0.27}$        &   91.16                                \\
             \quad~~~+ output                               & -      & $\text{87.70}_{\pm 2.24}$      & $\text{90.70}_{\pm 1.20}$        & $\text{92.19}_{\pm 0.11}$         & $\text{96.22}_{\pm 0.15}$         &   90.95                        \\
  \bottomrule
  
  \end{tabular}

    \caption{Performance of various dropping positions, structural patterns and compensation methods for RoBERTa-large model on RTE, MRPC, STS-B and SST-2 datasets. ``input'' and ``output'' refer to the dropout of input and output representations, respectively. The subscripts denote the standard deviation, while bold indicates the best performance. ``Compen.'' and ``Avg.'' are abbreviations for  compensation measures and the average results across four datasets.
        }
    \label{tab:4results}%
\end{table*}

\begin{figure*}[!t]
  \centering
  \includegraphics[width=\linewidth]{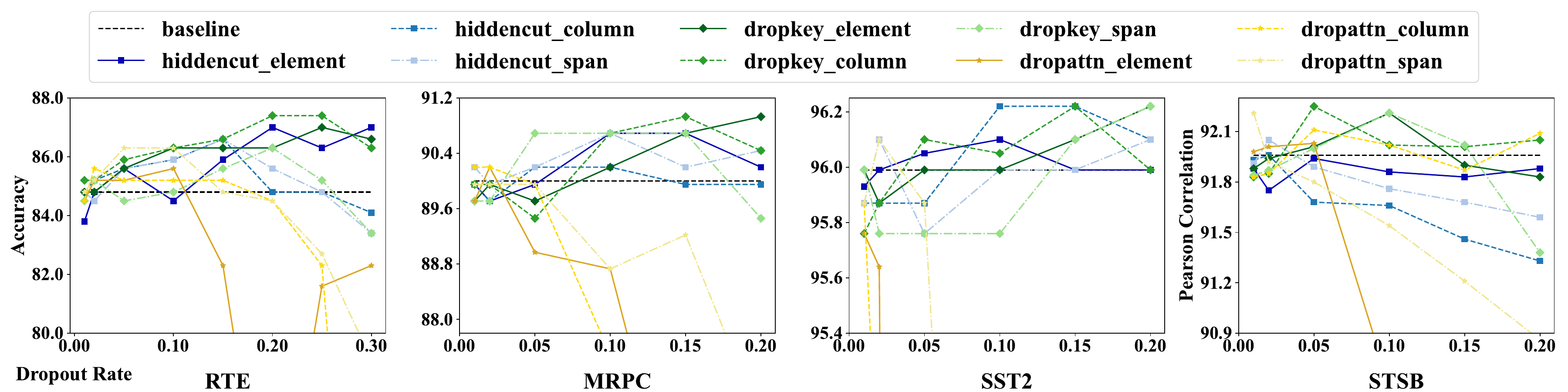}
  \caption{Performance of RoBERTa-large with different dropout methods on four NLU datasets, namely RTE, MRPC, SST-2 and STS-B. Markers and line styles differentiate various dropping positions, while the shades of color represent the structural patterns. Pearson correlation is reported for STS-B, while accuracy is utilized for others.}
  \label{fig:dropout results}
\end{figure*}

We first experiment with RoBERTa-large on four NLU datasets, and present the results in Table~\ref{tab:4results} and Figure~\ref{fig:dropout results}. 
Generally, almost all methods 
can outperform the baseline 
with a large margin. This demonstrates that despite limited trainable parameters, LoRA still suffers from overfitting and these transformer-specific dropout methods can alleviate this problem.
We claim that limited trainable parameters of LoRA still enable relatively large model capacity. This can stem from two aspects: (1) Even if the proportion is negligible, the number of tunable parameters remains significant due to the large size of foundation models. As mentioned earlier, there are still 0.79M tunable parameters, even if they only account for 0.22\% of the whole model. 
(2) Coupled with the base models, the expressiveness of these parameters is enlarged extremely, as evidenced by the remarkable performance 
in \citet{Hu2021}. 
This excessive model capacity contributes to the susceptibility to overfitting, despite only a negligible portion of trainable parameters.

\begin{table*}[!hb]
  \centering
\resizebox{0.85\textwidth}{!}{%
  \begin{tabular}{ccccccc}
  \toprule
  Model & Method & $\text{BLEU}\uparrow$ & $\text{NIST}\uparrow$ &                                        $\text{METEOR}\uparrow$ &                                         $\text{ROUGE\_L}\uparrow$ & $\text{CIDEr}\uparrow$ \\ \midrule
  \multirow{6}{*}{GPT2-Medium}  &     $\text{Full Finetuning}^{*}$ & $\text{68.20}$ & $\text{8.620}$ &                                        $\text{46.20}$ &                                         $\text{71.00}$ & $\text{2.470}$ \\ 
        &  Baseline & $\text{68.50}_{\pm{0.90}}$ & $\text{8.615}_{\pm{0.09}}$ &                                        $\text{46.43}_{\pm{0.26}}$ &                                         $\text{71.08}_{\pm{0.25}}$ & $\text{2.490}_{\pm{0.02}}$ \\
        &  HiddenCut & $\text{69.22}_{\pm{0.44}}$ & $\text{8.700}_{\pm{0.05}}$ &                                        $\text{46.66}_{\pm{0.11}}$ &                                         $\text{71.39}_{\pm{0.07}}$ & $\text{2.491}_{\pm{0.01}}$ \\
        &  DropKey & $\text{68.78}_{\pm{0.75}}$ & $\text{8.651}_{\pm{0.08}}$ &                                        $\text{46.53}_{\pm{0.24}}$ &                                         $\text{71.40}_{\pm{0.33}}$ & $\text{2.486}_{\pm{0.01}}$ \\
        &  \makecell{$\text{HiddenKey}^{-}$ } & $\text{69.35}_{\pm{0.48}}$ & $\text{8.726}_{\pm{0.04}}$ &                                        $\text{46.60}_{\pm{0.29 }}$ &                                         $\text{71.61}_{\pm{0.26}}$ & $\text{2.510}_{\pm{0.00}}$ \\
        &  HiddenKey & $\textbf{69.76}_{\pm{0.51}}$ & $\textbf{8.765}_{\pm{0.08}}$ &                                        $\textbf{46.80}_{\pm{0.11}}$ &                                         $\textbf{71.78}_{\pm{0.06}}$ & $\textbf{2.511}_{\pm{0.03}}$ \\ \midrule
  \multirow{2}{*}{ LLaMA2-7B}  &  Baseline & $\text{66.71}_{\pm{0.65}}$ & $\text{8.463}_{\pm{0.09}}$ & $\text{44.82}_{\pm{0.26}}$ & $\text{70.10}_{\pm{0.46}}$ & $\text{2.371}_{\pm{0.01}}$  \\
  &  HiddenKey & $\textbf{69.02}_{\pm{0.64}}$ & $\textbf{8.725}_{\pm{0.08}}$ & $\textbf{45.84}_{\pm{0.13}}$ & $\textbf{71.17}_{\pm{0.13}}$ & $\textbf{2.456}_{\pm{0.00}}$  \\ \bottomrule
\end{tabular}
}

\caption{Results of GPT2-Medium and LLaMA2-7B with various dropout methods on E2E NLG Challenge dataset.}
\label{tab:e2e}
\end{table*}

Different dropping positions prefer distinct structural patterns. 
As shown in Table~\ref{tab:4results}, the optimal structure for DropKey is ``column'', which deactivates specific keys across all queries within a head, thereby breaking the co-adaptation of value vectors and achieving better performance. Oppositely, \citet{Li2023} confirms the ineffectiveness of structural patterns in multiple CV tasks. This divergence may arise from 
that NLP tasks have a more semantically explicit token segmentation, while this property is absent for CV tasks. 
In comparison, HiddenCut only has one representation sequence instead of multiple ones in the multi-head self-attention module. Hence, ``column'' and ``span'' modes may erase too much information, especially when semantically important representations, such as emotional and negation ones, are masked. This could introduce excessive noise and even incorrect input-label pairs for more limited LoRA scenarios, 
and explains why element-wise HiddenCut achieves better performance on average, different from the span style for full finetuning \citep{Chen2021}. 

These dropout methods exhibit different characteristics in LoRA scenarios, and combining different positions can yield further improvement. Specifically, with a small dropout rate, all methods perform very similarly, fluctuating around the baseline. However, as the dropout rate increases, DropKey consistently achieves the best performance on four datasets, followed by HiddenCut. This might be partially attributed to the closer proximity of DropKey to LoRA.
In contrast, despite the similar dropping positions and the same forward pass as DropKey, DropAttention produces the worst results. This confirms our earlier analysis in Sec.~\ref{sec:math} that $\operatorname{NoGrad()}$ operator
leads to larger gradient noise in back-propagation and rapid   performance degradation as the dropout rate increases.
Considering their best performance, we further combine element-wise HiddenCut with column-wise DropKey, named $\text{HiddenKey}^{-}$. On average, it achieves additional improvement over any single dropout mechanism. We also attempt to combine DropAttention, but it does not result in any benefits.

As for the compensation measures to narrow the gap between training and inference stages, KL loss consistently achieves better performance than JS loss. Specifically, compared to $\text{HiddenKey}^{-}$ (i.e. HiddenKey without any additional loss), the introduction of KL loss always provides extra performance gains on all the datasets.
In contrast, JS loss does not have an apparent impact on the results, even if  \citet{Chen2021} claims its effectiveness in full finetuning settings.
This difference may arise from the more capacity-limited LoRA scenarios and superb dropout methods, which jointly squeeze the potential improvement space for augmented loss. Therefore, with the validated superiority, KL loss is incorporated into HiddenKey along the third dimension of our proposed framework. Due to the optimal practice along each dimension, HiddenKey steadily achieves the best performance among all the above methods and datasets.

\subsection{Complementarity with Input and Output Dropout}
In addition to DropKey, DropAttention and HiddenCut, which cover the transformer layer, 
cutoff is also applied to input embedding sequences for data augmentation \citep{Shen2020}, and standard dropout is used to the output representations for a more robust classifier. 
To comprehensively explore the impact of dropout on the entire model, we further investigate whether these methods could further enhance the transformer-specific dropout. 
The results at the end of Table~\ref{tab:4results} suggest that neither of these methods achieve consistent improvement over $\text{HiddenKey}^{-}$ across all the datasets, and both of their average performance suffers a slight decrease. This indicates that HiddenKey has predominantly captured the performance gains achieved through dropout methods, while dropping input or output does not contribute steady complementarity. This sufficiency hints that finetuning with HiddenKey only is enough in LoRA scenarios.

\subsection{Superiority on More NLU and NLG Tasks}

\begin{table*}[!ht]
  \centering
  \resizebox{\textwidth}{!}{%
  \begin{tabular}{cccccccccc}
  \toprule
  \multirow{2}{*}{Method} & \multicolumn{3}{c|}{A} & \multicolumn{3}{c|}{S} & \multicolumn{3}{c}{U} \\ \cmidrule(l){2-10} 
    & $\text{BLEU}\uparrow$ & $\text{METEOR}\uparrow$ & \multicolumn{1}{c|}{$\text{TER}\downarrow$} & $\text{BLEU}\uparrow$ & $\text{METEOR}\uparrow$ & \multicolumn{1}{c|}{$\text{TER}\downarrow$} & $\text{BLEU}\uparrow$ & $\text{METEOR}\uparrow$ & $\text{TER}\downarrow$ \\ \midrule
  $\text{Full Finetuning}^{*}$                 & $\text{46.50}$         & $\text{0.380}$ & $\text{0.530}$           & $\text{64.20}$          & $\text{0.450}$ & $\text{0.330}$           & $\text{27.70}$          & $\text{0.300}$ & $\text{0.760}$  \\ \midrule

  Baseline & $\text{54.78}_{\pm{0.16}}$ & $\text{0.411}_{\pm{0.00}}$ & $\text{0.395}_{\pm{0.00}}$ & $\text{62.30}_{\pm{0.47}}$ & $\text{0.420}_{\pm{0.04}}$ & $\text{0.331}_{\pm{0.00}}$ & $\text{45.53}_{\pm{0.21}}$ & $\text{0.376}_{\pm{0.00}}$ & $\text{0.464}_{\pm{0.00}}$ \\ 
HiddenCut & $\text{55.06}_{\pm{0.18}}$ & $\text{0.411}_{\pm{0.00}}$ & $\text{0.391}_{\pm{0.00}}$ & $\text{62.43}_{\pm{0.21}}$ & $\textbf{0.442}_{\pm{0.00}}$ & $\text{0.329}_{\pm{0.00}}$ & $\text{46.11}_{\pm{0.20}}$ & $\text{0.377}_{\pm{0.00}}$ & $\text{0.458}_{\pm{0.00}}$ \\
DropKey  & $\text{55.22}_{\pm{0.34}}$ & $\text{0.411}_{\pm{0.00}}$ & $\text{0.389}_{\pm{0.00}}$ & $\text{62.47}_{\pm{0.17}}$ & $\text{0.441}_{\pm{0.00}}$ & $\text{0.328}_{\pm{0.00}}$ & $\text{46.39}_{\pm{0.75}}$ & $\text{0.378}_{\pm{0.00}}$ & $\text{0.455}_{\pm{0.01}}$ \\ 
\makecell{$\text{HiddenKey}^{-}$ } & $\text{55.26}_{\pm{0.20}}$ & $\text{0.411}_{\pm{0.00}}$ & $\text{0.388}_{\pm{0.00}}$ & $\textbf{62.57}_{\pm{0.24}}$ & $\text{0.441}_{\pm{0.00}}$ & $\text{0.328}_{\pm{0.00}}$ & $\text{46.36}_{\pm{0.34}}$ & $\text{0.378}_{\pm{0.00}}$ & $\text{0.454}_{\pm{0.00}}$ \\ 
HiddenKey  & $\textbf{55.27}_{\pm{0.21}}$ & $\textbf{0.413}_{\pm{0.00}}$ & $\textbf{0.386}_{\pm{0.00}}$ & $\text{62.49}_{\pm{0.18}}$ & $\text{0.441}_{\pm{0.00}}$ & $\textbf{0.326}_{\pm{0.00}}$ & $\textbf{46.48}_{\pm{0.46}}$ & $\textbf{0.381}_{\pm{0.00}}$ & $\textbf{0.452}_{\pm{0.00}}$ \\
\bottomrule
  \end{tabular}%
  }

\caption{Results of GPT2-Medium finetuned with different dropout methods on WebNLG dataset. ``A'', ``S'' and ``U'' correspond to the ``All'', ``Seen'' and ``Unseen'' categories in the test set, respectively.}
\label{tab:webnlg}
\end{table*}

\begin{table}[!t]
  \centering
    \resizebox{0.35\textwidth}{!}{%
  
    \begin{tabular}{ccc}
    \toprule
    \multirow{2}{*}{Method}    & CoLA                            & QNLI            \\
    \hhline{|~--|}
                               & Matthew.           & Acc.            \\
    \midrule
    baseline                   & $\text{67.96}_{\pm 0.25}$       & $\text{94.23}_{\pm 0.17}$  \\
    HiddenKey                  & $\text{\bf 69.91}_{\pm 0.52}$       & $\text{\bf 95.04}_{\pm 0.11}$  \\
    \bottomrule
    \end{tabular}
    }

\caption{Results of RoBERTa-large finetuned with HiddenKey on CoLA and QNLI datasets.
  }
\label{tab: extra NLU}
\end{table}

\paragraph{More NLU Datasets.}
We further generalize 
HiddenKey to two extra NLU datasets, namely CoLA and QNLI. As shown in Table~\ref{tab: extra NLU}, HiddenKey steadily achieves 1.95 and 0.81 performance improvement over baselines on both of the datasets, 
reconfirming HiddenKey's superiority in NLU tasks.


\paragraph{NLG datasets.}
Following \citet{Hu2021}, we also experiment with GPT2-Medium 
on NLG tasks. 
As shown in Table~\ref{tab:e2e}, HiddenKey consistently outperforms full finetuning, LoRA baseline and other dropout methods over all the five metrics on E2E NLG Challenge dataset.
Similarly in Table~\ref{tab:webnlg}, on the ``All'', ``Seen'' and ``Unseen'' subsets of the WebNLG dataset, HiddenKey gains 7/9 wins over all other methods on BLEU, METEOR and TER metrics. 
Hence, HiddenKey exhibits a performance surge across diverse metrics, datasets and their subsets for NLG tasks, as it has shown for NLU tasks.


\subsection{Performance Boost on LLMs}
\begin{table}[!h]
  \centering
    \resizebox{0.35\textwidth}{!}{%
  
    \begin{tabular}{ccc}
    \toprule
    \multirow{2}{*}{Method}    & RTE                             & MRPC            \\
    \hhline{|~--|}
                               & Acc.                            & Acc.            \\
    \midrule
    baseline                   & $\text{88.45}_{\pm 0.79}$       & $\text{88.73}_{\pm 0.56}$  \\
    HiddenKey                  & $\text{\bf 90.25}_{\pm 1.05}$   & $\text{\bf 89.46}_{\pm 0.60}$  \\
    \bottomrule
    \end{tabular}
    }

    \caption{Results of LLaMA2-7B finetuned with HiddenKey on RTE and MRPC datasets. 
    \label{tab: llama nlu}
    }
\end{table}

With the dominance of LLMs, we also extend the application of HiddenKey to LLaMA2-7B, one of the most popular and open-sourced LLMs, on both NLU and NLG tasks. As shown in Table~\ref{tab: llama nlu}, models finetuned with HiddenKey outperform those without HiddenKey by a large margin on RTE and MRPC datasets. Similarly, HiddenKey consistently exhibits significant superiority on E2E NLG dataset across all metrics over baseline, shown at the end of Table~\ref{tab:e2e}. This indicates that HiddenKey can also function well with LLMs on diverse tasks.

\subsection{Ablation Study}
Based on our framework, we eliminate the components of HiddenKey to demonstrate the necessity of each dimension. As illustrated in Table~\ref{tab:4results}, \ref{tab:e2e} and \ref{tab:webnlg}, the substantial boost of $\text{HiddenKey}^{-}$ over previous methods and baselines on both NLU and NLG tasks indicates the significance of dropping positions and patterns in mitigating the susceptibility to overfitting in LoRA scenarios. Moreover, HiddenKey also consistently outperforms $\text{HiddenKey}^{-}$, emphasizing the importance of appropriate compensation measures. These results provide strong evidence for the effectiveness of our framework.

\section{Conclusion}
We investigate the possible contradiction between the limited trainable parameters of LoRA and overfitting associated excessive parameter redundancy. After confirming the overfitting-prone property of LoRA, we analyze existing dropout methods theoretically and empirically, and further introduce a unified framework for thorough comparison. This also guides us to derive a new dropout method, HiddenKey. With its superiority and sufficiency 
across multiple models and datasets, HiddenKey deserves to be the recommended dropout method to alleviate overfitting in LoRA-based scenarios.








\section{Limitation}
The main limitation of this work is the potentially longer training duration incurred by the Bidirectional Kullback-Leibler (KL) divergence loss. Specifically, the calculation of the KL loss requires the output distributions of two forward passes. 
In our implementation, as shown in Figure~\ref{fig:siamese}, we only perform back-propagation on one of the branches, resulting in approximately 50\% longer training time compared to the original training process. However, we argue that this can be greatly reduced by parallelizing the two forward passes with multiple processes. Alternatively, both branches can be back-propagated simultaneously or sequentially, before merging their gradient updates. This pipeline can be regarded as utilizing the same batch of samples twice, thereby roughly halving the number of iterations and resulting in similar total training time, which is left for future work.
Furthermore, it is worth noting that the training cost is one-time, and the introduction of KL loss can significantly improve models' performance, which is highly beneficial for performance-critical scenarios. On the other hand, for training cost-sensitive scenarios, using only $\text{HiddenKey}^{-}$ (i.e. HiddenKey without KL loss) can still outperform the baselines. Hence, we claim that despite the potential increase in training duration, HiddenKey and $\text{HiddenKey}^{-}$ do provide available options for different scenarios.

\section{Ethics Statement}
We strictly follow the ACL Code of Ethics during the research. To the best of our knowledge, there are no foreseeable potential risks in the methods we introduced. We report the computing infrastructure for all computational experiments presented in the paper. The transparent statistics on our results and detailed configuration of our experimental setup, including best-found hyperparameter values, are well stated. 
Besides, we will also release the code upon publication for publicly available reproducibility with minimal effort.


\section{Acknowledgement}
This work was supported in part by Hong Kong Innovation and Technology Support Programme Platform Research Project fund (ITS/269/22FP), the joint research scheme of the National Natural Science Foundation of China (NSFC) and Hong Kong Research Grants Council (RGC) (under grant N\_HKU714/21), and RGC grants 17204423 and C7004-22G (CRF).


\bibliography{arr2024_conference}

\newpage
\appendix
\section{Related work}

During the finetuning phase, full-finetuning involves updating all model parameters, resulting in a slightly modified version. However, with the rapid development of large language models (LLMs), this approach becomes increasingly impractical due to the high storage and inference expenses, particularly in multitask and personalized settings~\citep{wang2023multilora, chen2023netgpt}.
As lightweight alternatives, parameter-efficient finetuning (PEFT) methods only introduce or retrain a negligible portion of parameters, sharing most of the parameters while preserving competitive performance as full-finetuning \citep{Houlsby2019,Lester2021, Hu2021}. 
For instance, \citet{Houlsby2019} inserts and exclusively updates new adapters between pretrained layers, achieving remarkable performance with limited trainable parameters. However, this method increases the model's depth and incurs higher time latency. \citet{Lester2021} prefixes a learnable prompt to the input and feeds this longer sequence into the frozen model. Nevertheless, this approach reduces the available sequence length and is empirically shown to be sensitive to initialization. Similarly, \citet{Li2021} attaches prefixed tokens to the key and value sequences, addressing the first drawback but still suffering from the latter one. In contrast, BitFit~\citep{Zaken2021} only adjusts the biases,
effectively avoiding the aforementioned problems. However, its limited capacity leads to inferior performance. More recently, LoRA~\citep{Hu2021} imposes a low-rank decomposition on weight updates, which can be optionally merged into the original weights during inference, and avoids all the aforementioned issues.

Dropout~\citep{Hinton2012} randomly deactivates each neuron with a specific probability during training,
which can prevent the co-adaptation of neurons and has been extended to improve the performance of transformer models \citep{Zehui2019, Chen2021, Li2023}.
Specifically, \citet{Zehui2019} proposes the first variant specially designed for self-attention mechanism, DropAttention, which drops the attention weights randomly and applies normalized rescaling to ensure their sum to be one. 
Instead, HiddenCut~\citep{Chen2021} applies contiguous span-style masks to hidden representations in the feed-forward module. Recently, \citet{Li2023} introduces a drop-before-softmax scheme, HiddenKey, which drops key units before the softmax layer so that the sum of attention weights can be kept as one automatically. However, it only focuses on computer vision tasks, while totally neglecting NLP tasks that emphasizes semantics and linguistic information.
During inference, dropout is usually disabled by default for better performance and deterministic outputs. 
However, this is not consistent with the training stage and can lead to a gap between the actual and ideal performance. In order to address this divergence, 
R-Drop~\citep{Wu2021} minimizes the bidirectional Kullback-Leibler divergence between the output distributions of two forward passes with dropout for more noise-resilient outputs. In comparison, \citet{Shen2020} narrows this gap by applying Jensen-Shannon Divergence loss to enforce consistent representations between outputs with and without dropout.

\section{Overfitting-Prone Property of LoRA}
As an illustrative example, Figure~\ref{fig:dropout} shows the evaluation accuracy of LoRA with different ranks on the RTE dataset. This clearly indicates that with the increase of the rank and trainable parameters, the performance of LoRA initially improves and then deteriorates due to progressively excessive parameter redundancy, demonstrating the susceptibility to overfitting in LoRA scenarios.
\begin{center}
    \includegraphics[width=0.95\linewidth]{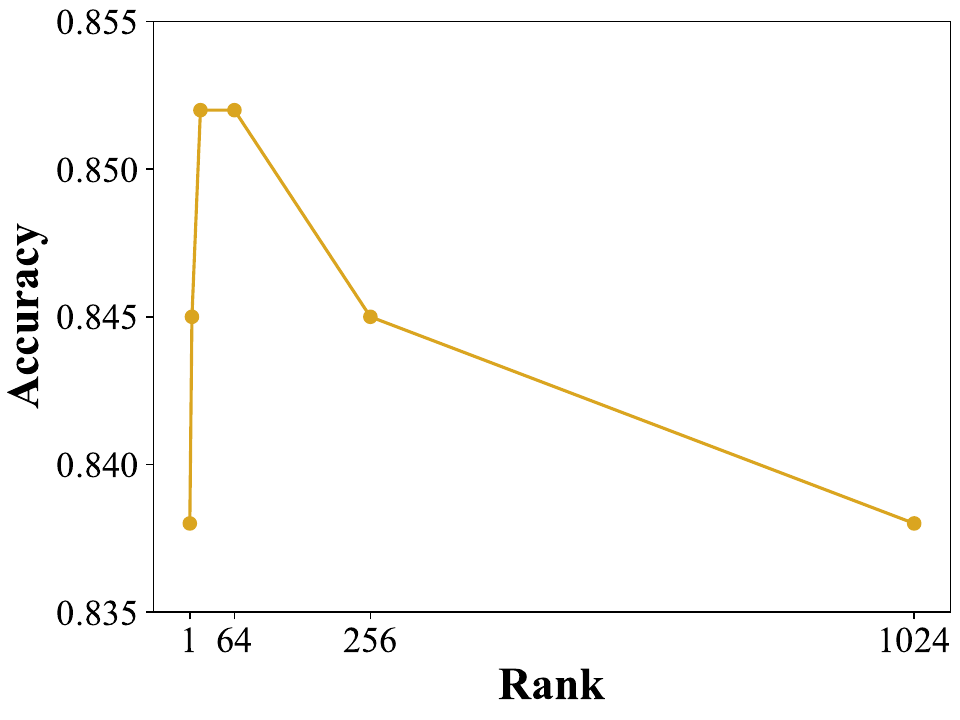}
    \captionof{figure}{Evaluation accuracy of LoRA with respect to the rank on RTE dataset.}
    \label{fig:dropout}
\end{center}

\section{Mathematical Proofs} \label{apdx: math}
We prove the mathematical equivalence of $w_u^\prime$ for DropKey and DropAttention as follows: 

{\small
\begin{align}
    & \frac{\exp (g_u)}{\sum_{i=0}^{l-1} \exp (g_i)} \cdot \frac{1}{\sum_{i=0}^{l-1}{\overline{w}_i}} \displaybreak[0] \notag \\
    = &\frac{\exp (g_u)}{\sum_{i=0}^{l-1} \exp (g_i)} \cdot \frac{1}{1-{w}_m} \displaybreak[0] \notag \\
    = &\frac{\exp (g_u)}{\sum_{i=0}^{l-1} \exp (g_i)} \cdot \frac{1}{1-\frac{\exp (g_m)}{\sum_{i=0}^{l-1} \exp (g_i)}} \displaybreak[0]  \label{eq:equivalence} \\
    = & \frac{\exp (g_u)}{\sum_{i=0}^{l-1} \exp (g_i) - \exp (g_m)} \displaybreak[0] \notag \\
    = & \frac{\exp (g_u)}{\sum_{i=0, \neq m}^{l-1} \exp (g_i) } \displaybreak[0] \notag \\
    = & \frac{\exp (g_u^\prime)}{\sum_{i=0}^{l-1} \exp (g_i^\prime)} \notag    
\end{align}
}
The proportional relationship of $\frac{\partial w_u^\prime}{\partial g_u}$ between DropKey and DropAttention can be derived with the following equation:


\begin{small}
\begin{align}
&\frac{(\frac{\partial w_u^\prime}{\partial g_u} )^{\text{DropKey}}} {(\frac{\partial w_u^\prime}{\partial g_u} )^{\text{DropAttention}}} \nonumber \\
= & \frac{ \exp(g_u) \cdot  (\sum_{i=0, \neq m}^{l-1} \exp (g_i) - \exp(g_u))}{ (\sum_{i=0, \neq m}^{l-1} \exp (g_i))^2} \label{eq:k prove}  \\ 
& \cdot \frac{\sum_{i=0}^{l-1} \exp (g_i) \cdot \sum_{i=0, \neq m}^{l-1} \exp (g_i)}{\exp(g_u) \cdot \sum_{i=0, \neq u}^{l-1} \exp (g_i) } \nonumber \\ 
= & \frac{\sum_{i=0, \neq m}^{l-1} \exp (g_i) - \exp(g_u)}{\sum_{i=0, \neq m}^{l-1} \exp (g_i)}  \nonumber \\
& \cdot \frac{\sum_{i=0}^{l-1} \exp (g_i)}{\sum_{i=0, \neq u}^{l-1} \exp (g_i) }  \nonumber \\
= & \frac{1-\frac{\exp(g_u)}{\sum_{i=0, \neq m}^{l-1} \exp (g_i)}}{1-\frac{\exp(g_u)}{\sum_{i=0}^{l-1} \exp (g_i)}} \nonumber
\end{align}
\end{small} \\
Denoting $k$ as the result of Eq.~\ref{eq:k prove}, we have 
{\small
\begin{align}
 k &< \frac{1-\frac{\exp(g_u)}{\sum_{i=0, \neq m}^{l-1} \exp(g_i) + \exp(g_m)}}{1-\frac{\exp(g_u)}{\sum_{i=0}^{l-1} \exp (g_i)}} \\
  &= 1 \nonumber
\end{align}}

\section{Dataset Details} \label{apdx: dataset}

For NLU tasks,
(i) Stanford Sentiment Treebank (\textbf{SST-2}) \citep{Socher_Perelygin_Wu_Chuang_Manning_Ng_Potts_2013} is an English sentiment classification benchmark for a single sentence task, predicting whether the sentiment of movie reviews is positive or not.
(ii) Recognizing Textual Entailment (\textbf{RTE}) \citep{Wang2018} presents an inference task that predicts the entailment relation between two sentences.
(iii) Microsoft Research Paraphrase Corpus (\textbf{MRPC}) \citep{Dolan_Brockett_2005} predicts the semantic equivalence between two sentences, while 
(iv) Semantic Textual Similarity Benchmark (\textbf{STS-B}) \citep{Cer2017} predicts the similarity between two sentences. The later two tasks are involved with comparing and assessing the similarity and paraphrasing of two sentences. 
Notably, compared to the other classification tasks, STS-B performs a regression task and thus encompasses a broad range of tasks, enhancing the generalizability of our conclusions.
Besides, additional experiments are further conducted
on 
(v) Corpus of Linguistic Acceptability (\textbf{CoLA}) \citep{Warstadt2018}, which aims to predict whether a sentence is linguistically acceptable or not, and 
(vi) Question Natural Language Inference (\textbf{QNLI}) \citep{Rajpurkar2018}, which predicts whether a sentence is the answer to a given question.
For NLG tasks, we focus on 
(vii) \textbf{E2E} NLG Challenge \citep{Novikova2017} and 
(viii) \textbf{WebNLG} \citep{Gardent_Shimorina_Narayan_Perez-Beltrachini_2017}. The former consists of sets of slot-value pairs along with multiple corresponding natural language references in the restaurant domain, while the later is a dataset where models generate the corresponding description in form of natural language text given a sequence of SUBJECT-PROPERTY-OBJECT triples.

As for the evaluation metrics, we report the Pearson correlation for STS-B, 
Matthew's correlation for CoLA, and accuracy for other NLU datasets. 
For NLG tasks, BLEU, NIST, METEOR, ROUGE-L and CIDEr are used on the E2E NLG Challenge dataset, while BLEU, METEOR and TER are evaluated separately for ``Unseen'', ``Seen'' and ``All'' categories in the test set of the WebNLG dataset. 

\section{Hyperparameter Configuration} \label{apdx: config}
\begin{table*}[!ht]
    \centering
    \begin{tabular}{c|cccccc|cc}
    \toprule
    Model                       & \multicolumn{6}{c|}{RoBERTa-large}   & \multicolumn{2}{c}{ LLaMA2-7B} \\ \midrule
    Dataset                     & RTE     & MRPC    & STS-B    & SST-2      & CoLA    & QNLI        & RTE     & MRPC   \\ \midrule
    Optimizer                   & \multicolumn{6}{c|}{AdamW}                & \multicolumn{2}{c}{AdamW} \\
    Weight Decay                & \multicolumn{6}{c|}{0.1}                  & \multicolumn{2}{c}{0.1} \\
    Warmup Ratio                & \multicolumn{6}{c|}{0.06}                 & \multicolumn{2}{c}{0.06} \\
    LR Schedule                 & \multicolumn{6}{c|}{Linear}               & \multicolumn{2}{c}{Linear} \\
    Learning Rate               & 4E-4    & 3E-4    & 3E-4     & 4E-4       & 2E-4    & 2E-4        & \multicolumn{2}{c}{5E-4}     \\
    Epoch                       & 30      & 30      & 10       & 10         & 40      & 10          & 10      & 8    \\
    Batch Size                  & 64      & 32      & 32       & 64         & 32      & 32          & 64      & 32     \\
    Mac Seq. Len.               & 512     & 512     & 128      & 512        & 128     & 512         & \multicolumn{2}{c}{512}     \\
    LoRA Rank                   & \multicolumn{6}{c|}{$r_q=r_v=8$}          & \multicolumn{2}{c}{$r_q=r_v=8$} \\
    LoRA Scalar                 & \multicolumn{6}{c|}{16}                   & \multicolumn{2}{c}{16} \\ \bottomrule
    \end{tabular}%
    \caption{Hyperparameters for RoBERTa-large and LLaMA2-7B models with LoRA on NLU datasets.}
    \label{tab: nlu hyper}
\end{table*}
    
\begin{table}[!ht]
    \centering
    \resizebox{\linewidth}{!}{
    \begin{tabular}{c|cc}
    \toprule
    Dataset                     & E2E NLG Challenge                                         & WebNLG \\ 
    \midrule
                            \multicolumn{3}{c}{Training} \\ 
    \midrule
    Optimizer                   & \multicolumn{2}{c}{AdamW} \\
    Weight Decay                & \multicolumn{2}{c}{0.01} \\
    Warmup Step                 & \multicolumn{2}{c}{500} \\
    LR Schedule                 & \multicolumn{2}{c}{Linear} \\
    Learning Rate               & \multicolumn{2}{c}{2E-4} \\
    Epoch                       & \multicolumn{2}{c}{5} \\
    Batch Size                  & \multicolumn{2}{c}{8} \\
    Label Smooth                & \multicolumn{2}{c}{0.1} \\
    LoRA Rank                   & \multicolumn{2}{c}{$r_q=r_v=4$} \\
    LoRA Scalar                 & \multicolumn{2}{c}{32} \\ 
    \midrule
                            \multicolumn{3}{c}{Inference} \\ 
    \midrule
    Beam Size                   & \multicolumn{2}{c}{10} \\
    Length Penalty              & 0.9                                         & 0.8 \\
    No Repeat N-Gram Size        & \multicolumn{2}{c}{4} \\
    Repetition Penalty          & \multicolumn{2}{c}{1.0} \\ 
    \bottomrule
    \end{tabular}%
    }

    \caption{Hyperparameters for GPT2-Medium with LoRA on NLG datasets.}
    \label{tab: nlg hyper}
\end{table}

As shown in Table~\ref{tab: nlu hyper} and \ref{tab: nlg hyper}, we mainly follow the setup of LoRA \citep{Hu2021} with as minimal changes as possible. 
However, based on our pre-experiments, significant fluctuations of the results are observed when models are trained with the original epochs, even if only random seeds change. Therefore, we increase the number of training epochs for more steady results. We also use the regular initialization instead of the MNLI checkpoint for LoRA modules. Different from RoBERTa-large and GPT2-Medium models, we employ FP16 mixed precision training for LLaMA2-7B to reduce the memory consumption, and set the epoch to one. Besides, we utilize greedy search with length penalty of 1.0 and ``no repeat n-gram size'' of 0 for inference, which empirically outperforms the settings of GPT2-Medium.

For the specific parameters in our experiments, we disable dropout in baselines and iterate all available dropout rate from \{0.01, 0.02, 0.05, 0.1, 0.15, 0.2\} for various dropout methods, which is expanded with \{0.25, 0.3\} for clearer trend of performance in RTE dataset. 
To the best of our knowledge, neither of HiddenCut, DropKey and DropAttention implements experiments with a casual decoder-only transformer model before. Based on our empirical observation, applying any of these methods can only produce limited improvement or even degradation on both NLU and NLG tasks, and the results are extremely sensitive to the dropout rate. This phenomenon might be caused by fragile shallow forwarding pass. In other words, noise introduced by dropout methods can be amplified with the propagation and diminish the benefits brought by dropout. Hence, we only introduce the dropping in the latter half of layers in decoder-only models and the apparent performance improvement emerges again. 
Besides, our pre-experiments demonstrate that a weight between 0.01 and 10 for KL and JS loss generally yields the best results. Therefore, we iterate the weight within \{0.01, 0.02, 0.05, 0.1, 0.2, 0.5, 1, 2, 5, 10\}.
All experiments are repeated 5 times on a NVIDIA V100 GPU to calculate the median values for NLU tasks, while the average values of three runs on a NVIDIA A100 GPU is reported for NLG tasks.

\section{Finetuning dynamics} 
\begin{figure}[h]
  \includegraphics[width=\linewidth]{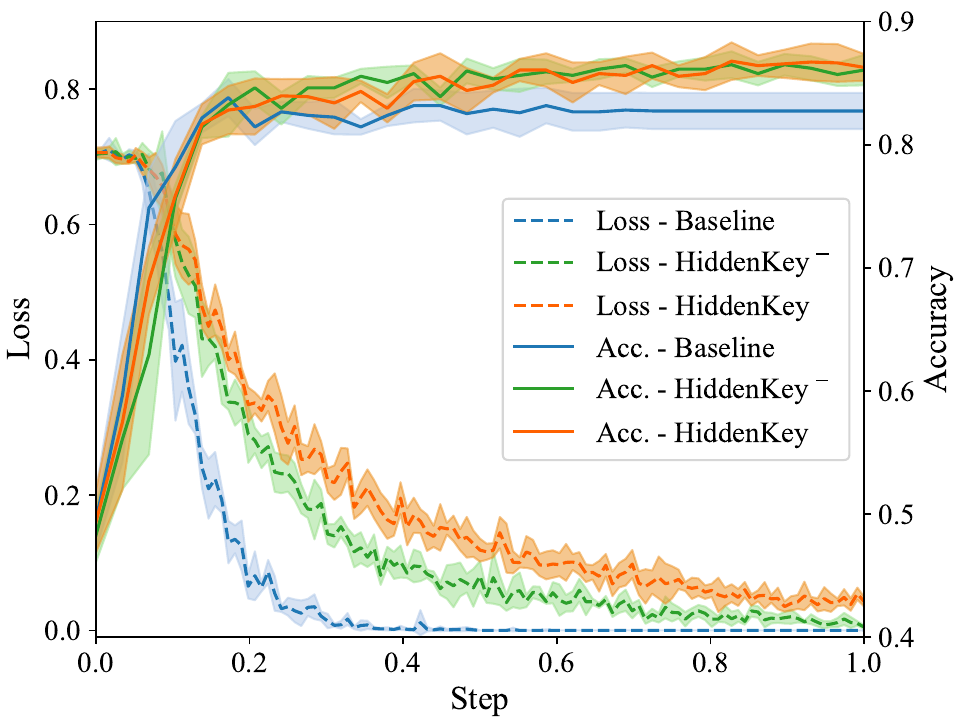}
  \caption{Finetuning loss and evaluation accuracy for baseline, $\text{HiddenKey}^{-}$ and HiddenKey. 
  }
  \label{fig:dynamic curves}
\end{figure}

Beyond the superior performance of HiddenKey, we also visualize the finetuning dynamics for a deeper understanding. 
Figure~\ref{fig:dynamic curves} presents the average dynamic curves of training loss and evaluation accuracy across five random seeds for multiple methods on the RTE dataset.
Compared to the baseline whose training loss rapidly converges to near zero, the introduction of $\text{HiddenKey}^{-}$ (i.e. column-wise DropKey and element-wise HiddenCut) slows down this process and leads to larger final loss.
However, large final loss does not mean inferior performance. 
Specifically, after reaching a fair peak value,
accuracy of the baseline deteriorates with the continuous loss decline. This hints that the models suffer from overfitting, which further supports our earlier analysis. In contrast, $\text{HiddenKey}^{-}$ reaches the peak accuracy slightly slowly
but remains superior to the baseline. With the additional KL loss, the accuracy keeps fluctuating upwards and achieves the best performance.
It can be anticipated that a longer finetuning process would result in higher accuracy for HiddenKey. 
In summary, LoRA-based PEFT scenarios are still overfitting-prone, while HiddenKey can provide excellent model regularization in such settings, 
and continues improving the performance when further finetuning is allowed.

\section{Statistical Significance Test}
\begin{table}[h]
\resizebox{0.9\linewidth}{!}{
\begin{tabular}{@{}ccc@{}}
\toprule
Model (Benchmark)           & Method    & p-value \\ \midrule
\multicolumn{1}{c|}{}        & Baseline   & 0.080   \\ \cmidrule(l){2-3} 
\multicolumn{1}{c|}{RoBERTa} & DropKey    & 0.059   \\ \cmidrule(l){2-3} 
\multicolumn{1}{c|}{(GLUE)}  & HiddenCut  & 0.024   \\ \cmidrule(l){2-3} 
\multicolumn{1}{c|}{}        & HiddenKey- & 0.031   \\ \midrule
\multicolumn{1}{c|}{}        & Baseline   & 0.044   \\ \cmidrule(l){2-3} 
\multicolumn{1}{c|}{GPT-2}   & DropKey    & 0.052   \\ \cmidrule(l){2-3} 
\multicolumn{1}{c|}{(E2E)}   & HiddenCut  & 0.041   \\ \cmidrule(l){2-3} 
\multicolumn{1}{c|}{}        & HiddenKey- & 0.043   \\ \bottomrule
\end{tabular}
}%
\caption{P-values of HiddenKey versus alternative methods.}
\label{tab: p-test}
\end{table}

To assess the statistical significance of the results presented in Table~\ref{tab:4results} and Table~\ref{tab:e2e}, we calculate the p-values of HiddenKey comparing against alternative approaches, averaged on the benchmarks. As shown in Table~\ref{tab: p-test}, the obtained p-values, all below 0.1 with the majority falling below 0.05, strongly indicate the statistical significance of HiddenKey's superiority.

\end{document}